\documentclass[conference]{article}
\usepackage{ijcai18}
\usepackage{times}
\usepackage{helvet}
\usepackage{courier}
\usepackage{xcolor}
\usepackage{soul}
\usepackage[utf8]{inputenc}
\usepackage[small]{caption}
\usepackage{verbatim}
\usepackage{subfigure}
\usepackage{color}
\usepackage{float}
\usepackage{array}
\usepackage{booktabs}
\usepackage{graphicx}
\usepackage{amsmath}
\usepackage{amssymb}
\usepackage{amsfonts}
\usepackage{algorithm}
\usepackage{algpseudocode}
\usepackage{balance}
\usepackage{tabu}
\usepackage{multirow}
\usepackage[flushleft]{threeparttable}
\PassOptionsToPackage{hyphens}{url}
\usepackage{hyperref}
\usepackage{bbold}

\newcommand{\hide}[1]{} 
\newcommand{\vpara}[1]{\vspace{0.1in}\noindent\textbf{#1 }}

\newcommand{\figref}[1]{Figure~\ref{#1}} 
\newcommand{\beq}[1]{\begin{equation}#1\end{equation}}

\begin{document}

\title{
Weakly Learning to Match Experts in Online Community}

\author{Yujie Qian$^{\dagger \ddagger}$\thanks{Work performed while the author was at Tsinghua University.} \quad Jie Tang$^\dagger$ \quad Kan Wu$^\dagger$\\
$^\dagger$Tsinghua University \quad
$^\ddagger$Massachusetts Institute of Technology   \\
\href{mailto:yujieq@csail.mit.edu}{yujieq@csail.mit.edu} ~~~ 
\href{mailto:jietang@tsinghua.edu.cn}{jietang@tsinghua.edu.cn} ~~~  
\href{mailto:wu-k14@mails.tsinghua.edu.cn}{wu-k14@mails.tsinghua.edu.cn}
}


\maketitle

\begin{abstract}
In online question-and-answer (QA) websites like Quora, one central issue is to find (invite) users  who are able to provide answers to a given question and at the same time would be unlikely to say ``no'' to the invitation. 
The  challenge is how to trade off the matching degree between users' expertise and the question topic, and the likelihood of positive response from the invited users.
In this paper, we formally formulate the problem and develop a weakly supervised factor graph (WeakFG) model to address the problem. The model explicitly captures expertise matching degree between questions and users. 
To model the likelihood that an invited user is willing to answer a specific question, 
we incorporate a set of correlations based on  social identity theory into the WeakFG model.
We use two different genres of datasets: QA-Expert and Paper-Reviewer, to validate the proposed model. 
Our experimental results show that the proposed model 
can significantly outperform (+1.5-10.7\% by MAP)
the state-of-the-art algorithms for matching users (experts) with community questions.
We  have also developed an online system 
to further demonstrate the advantages of the proposed method.
\end{abstract}
\section{Introduction}
\label{sec:intro}

Led by technology innovation, the global economy is in transition to a ``knowledge economy''. It is common to post a question on  knowledge sharing or question-and-answer websites such as Quora
and invite  ``experts'' to help answer the question. 
Thus, finding appropriate experts for 
the given question becomes a central task. 
In the past, related research including  experts finding for
 community-based questions~\cite{riahi2012finding,zhao2015expert},  peer reviewer recommendations for 
 papers~\cite{deng2008formal,djupe2015peer,t.dumais:automating,goldsmith2007ai,long2013good,Tang:11Neuro}, and expertise modeling~\cite{han2016distributed,Mimno:07,Zhang:07WWW} has been conducted. A  survey can be also found in~\cite{Balog:12Trends}. 
 However, 
 most existing research does not consider the second half of the problem --- the probability  of
 whether the invited experts would accept to answer the question.
Indeed, the subproblem is even more serious---e.g. more than half of the questions on Quora only have one or do not have any answer\footnote{\url{https://www.quora.com/What-percentage-of-questions-on-Quora-have-no-answers}}.

\hide{
For example, Quora, Zhihu, and TouTiao Q\&A
have attracted big-time experts in numerous fields to answer various questions. 
Another example is peer review, an important part of scientific publishing~\cite{rennie1999editorial}. 
Many researchers consider peer review as part of their professional responsibilities. Publishing groups also use the quality of peer reviews as an indicator of the success of a journal. 
At a high level, the problem is referred to as \textit{expertise matching} --- matching experts with questions.
However, existing solutions for expertise matching are far from satisfactory. 
One challenge of this problem is how to find appropriate experts who are qualified to answer the corresponding questions. 
The other one is how to find experts who will agree to provide their answers.
One would expect that experts with sufficient knowledge would be the best to answer a given question. 
But in practice, the latter is even more serious.
Statistics show that about half of the questions on Quora only have one or do not have any answer\footnote{https://www.quora.com/What-percentage-of-questions-on-Quora-have-no-answers}.
}

These QA websites such as Quora 
are now striving to match questions with the \textit{right} experts and seek the best respondent to provide an answer.
The problem can be formalized as follows: 
Given a query (i.e., question/document) $q$, one of the goals is to find experts with sufficient knowledge.
We consider a set of $N$ candidate experts $E=\{e_1, \cdots, e_N\}$. 
We use function $R(q, e_i)$ to measure the extent to which expert $e_i$ has the knowledge on query $q$.
The other goal here is to estimate how likely expert $e_i$ would accept to answer the query $q$. We use function $R'(q, e_i)$ to quantify the probability of acceptance. Finally, the above problem can be defined as: given a query $q$ and a set of experts $E=\{e_1, \cdots, e_N\}$, the goal is to maximize the utility function:
\beq{
	S_{qi} = \alpha R(q, e_i) + (1-\alpha) R'(q, e_i) \label{eq:problem}
}

\noindent where $\alpha$ is a tunable parameter to trade off the importance between the matching degree of expert $e_i$ with query $q$ and the acceptance probability of expert $e_i$ to answer $q$. Setting $\alpha=1$ would reduce the problem to the traditional expert finding problem without considering the acceptance probability of the invited expert.

The problem defined in Eq. (\ref{eq:problem}) is much  more challenging than  the traditional expert finding problem.
This is because factors that affect an expert to accept (or reject) an answer invitation are very complicated.
For example, it might be because the expert is overloaded or simply due to
the unpopularity of the topic.
Moreover, how to instantiate the function $R'(q, e_i)$ in oder to quantify the acceptance probability is another challenge. From the algorithm perspective, we can feed some training data to a machine learning algorithm to learn a predictive model; however, the trouble is how to collect sufficiently labeled data.
The last challenge is how to evaluate the performance of a potential solution, in particular in an online fashion. 

\hide{
For peer review, other reasons for experts (researchers) to decline a review invitation include having too many reviews at hand and a tight deadline for completing the review \cite{tite2007peer,willis2016peer}. Another survey of political science journals also finds that decline rate is related to the expert's personal experience, e.g., top experts are more likely to decline than junior experts~\cite{djupe2015peer}. 
Technically, the challenge is how to design a principled approach to deal with the problem of expertise matching by considering the expert response.
}

\vpara{Our Solution and Contributions.}
In this paper, we formalize the problem into a weakly supervised factor graph (WeakFG) model to address the aforementioned challenges.
We introduce embedding-based metrics to  measure the matching degree between  queries and experts.
WeakFG is  able to learn from response logs to predict who are more likely to accept/decline an answer invitation. To  address the problem of lacking labeled data, WeakFG defines a set of correlation functions and uses the correlations to propagate the labeled information to unlabeled data.

\hide{
combining them with other features and correlations in the ranking model.
RankFG is able to learn from history response log and predict who is at a high risk of declining an answer invitation. 
}

To validate the effectiveness of the 
proposed model, we conduct both \textit{offline} and \textit{online} evaluations. 
For the offline evaluation, we first use the data from an international competition in 2016\footnote{\url{https://biendata.com/competition/bytecup2016/}}. The task is to predict the probability that an expert answers a given question in online communities.
Experimental results show that the proposed model can achieve  better performance (1.5-10.7\% by MAP) than several state-of-the-art algorithms.
To validate the generality of the proposed model, 
we applied it to a related, but somehow different, setting: peer-reviewer recommendation --- recommending reviewers to scientific articles.
We use the bidding information of a major computer science conference with $\sim1,000$ submissions and $440$ program committee members  to evaluate the prediction on acceptance of review invitation. We achieve the same better performance than several comparison methods.
For online evaluation, we developed a Chrome Extension of reviewer recommendation and deployed it in the Google Chrome Web Store. About 50 journal editors downloaded the extension and used it for reviewer recommendations.
Feedbacks from users also demonstrate the effectiveness of the proposed method.

\section{Related Work}
\label{sec:related}

Related research 
can be traced back to 30 years ago. 
We briefly survey existing literature from three aspects: expert finding, expert matching, and expertise modeling.

Expert finding aims to identify experts for given topic/query. Applications include finding experts to answer question in online communities~\cite{liu2005finding,riahi2012finding,zhao2015expert}, finding experts to review scientific articles~\cite{yang2009reviewer,djupe2015peer,Haym:99IJCAI}, finding experts from bibliography data~\cite{deng2008formal,zhang2008mixture}, or finding software developers to fix a bug~\cite{xia2017improving}. 
This category of research is closely related to our work. Those methods can be used to instantiate the left term in Eq. (\ref{eq:problem}).
The second category of related research is expert matching. The goal is to assign a list of experts to a list of queries by considering  matching degree, load balance, and  other constraints. There was a bunch of research on this topic, probably starting from conference-paper-reviewer assignment~\cite{Benferhat:01,t.dumais:automating,goldsmith2007ai,Karimzadehgan:09,kou2015weighted,long2013good,Tang:11Neuro} to recently studies for online community and crowdsourcing~\cite{hung2015minimizing}.
One problem which has long been criticized~\cite{ishiyama2014annual,ishiyama2015report} in expert matching is that it does not consider the expert's willingness likelihood, i.e., the second term in Eq. (\ref{eq:problem}).
The study of expertise modeling focuses on mining experts' interests or expertise. Methods using probabilistic model~\cite{Mimno:07}, network analysis~\cite{Zhang:07WWW}, and embedding~\cite{han2016distributed} have been developed. However, they  did not consider the experts' acceptance probability.

\hide{
Probabilistic models try to improve the expertise matching accuracy between experts and papers. Early works such as latent semantic indexing \cite{t.dumais:automating} and  keyword matching \cite{Haym:99IJCAI} tried to solve the problem in an information retrieval system. Later, Mimno~\textit{et al.} compared several language models and proposed Author-Persona-Topic model \cite{Mimno:07}. Karimzadehgan~\textit{et al.} proposed reviewer aspect modeling and paper aspect modeling, and tried to model multiple aspects of expertise \cite{Karimzadehgan:08}. Han~\textit{et al.}  studied learning expertise representations in collaborative networks \cite{han2016distributed}.

Optimization model concentrates on solving the optimization problem of constructing panels between a list of reviewers and a list of papers. These models usually consider the constraints in conference paper-reviewer assignment tasks \cite{Benferhat:01}, such as conflict of interests, paper demand and reviewer workload. Many methods have been proposed, such as search and greedy algorithm \cite{long2013good,kou2015weighted}, integer linear programming \cite{Karimzadehgan:09}, network flow \cite{goldsmith2007ai} and minimum cost flow~\cite{Tang:11Neuro}. 
}

A few systems have  been developed to make reviewer recommendations, e.g., AMiner~\cite{Tang:08KDD}.
Toutiao  organized a data challenge to predict the probability that an expert would accept the invitation to answer specific questions. 

\hide{
Wu \textit{et al.}  presents a patent partner recommendation framework in enterprise social networks, which also uses a ranking factor graph model \cite{wu2013patent}. 
}



\begin{figure}[t]
	\centering
	\mbox{
		\includegraphics[height=1.55in]{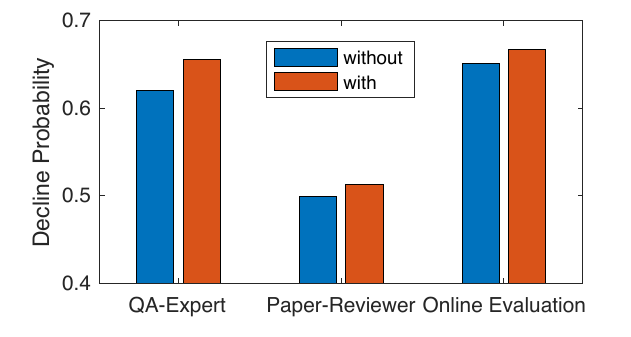}
	}
	\vspace{-0.17in} 
	\caption{
		Decline probability of an expert conditioned on whether or not the expert has a correlated ``friend'' who has already declined on two datasets: QA-Expert and Paper-Reviewer, and our online evaluation of journal reviewer recommendation. \label{fig:delinecorr}}
	\vspace{-0.05in}
\end{figure}

\section{Methodologies}
\label{sec:approach} 

Given a query (e.g., question/paper) $q$, our goal is to find experts with sufficient knowledge 
who are willing to answer the question or review the paper, with limited amount of labeled data.
Let us consider a set of $N$ candidate experts $E=\{e_1, \cdots, e_N\}$. The general problem can be then formalized as seeking a ranking function to quantify each expert $e_i$'s expertise degree and acceptance probability simultaneously.

The ranking score $S_{qi}$ was defined by only considering the expertise matching degree between $q$ and the expert $e_i$ (Cf. the first term in Eq. (\ref{eq:problem})).
One method
is to use language models~\cite{Zhai:01SIGIR}, which estimates the generative probability of a query given each document (expert-related document here) using a smoothed word distribution. 
Afterwards, topic models 
have been used to model latent  topics in order to improve the  matching accuracy~\cite{Mimno:07,riahi2012finding,Tang:08KDD}. 

However, all these models do not consider the  probability that  expert $e_i$ accepts to answer the query $q$. 
Algorithmically,  the difficulty is how to define the second term in Eq. (\ref{eq:problem}). 
Another challenge, in practical scenarios,  is the lack of sufficient labeled data.
One potential solution to this end is  weakly supervised learning~\cite{crandall2006weakly}, which essentially leverages some kind of correlation between training instances to propagate the labeled information to unlabeled instances. Widely used correlations include similarity and links (e.g., friend or coauthor relationships).
Figure~\ref{fig:delinecorr} shows the decline probability of an expert, conditioned on whether or not the expert has a  ``friend'' who has already declined. We see correlation indeed exists  in all the datasets used in our experiments (Cf. \S 4 for details of the datasets).
Now, finding useful correlations becomes an important issue. Social relationships like friendships can be considered as correlations, but are still sparse.
We present an interesting idea of defining correlations using
social identify theory~\cite{turner1986significance,tajfel2004social}, 
which suggests that people with the same social identity (e.g., position, affiliation, age, etc.) tend to share the same social behavior. 
For example, in an online community, users with the same job
tend to accept/decline the same question invitation; in the paper-reviewer recommendation setting, researchers from the same organization are more likely to decline an invitation at the same time.
We formalize the social identify-based correlations into a weakly supervised factor graph (WeakFG) model, to simultaneously learn the expertise matching degree and the acceptance probability with limited labeled data.

\subsection{WeakFG: Weakly Supervised Factor Graph}
We now explain how WeakFG works.
In WeakFG, we first use the embedding technologies~\cite{mikolov2013distributed} to capture  \textit{semantics} of expertise matching
and then use a semi-supervised graph model to predict whether expert $e_i$ will accept or decline an invitation, by considering both local similarity and correlation between experts.

\begin{figure}[t]
	\centering
	\mbox{
		\includegraphics[height=2.05in]{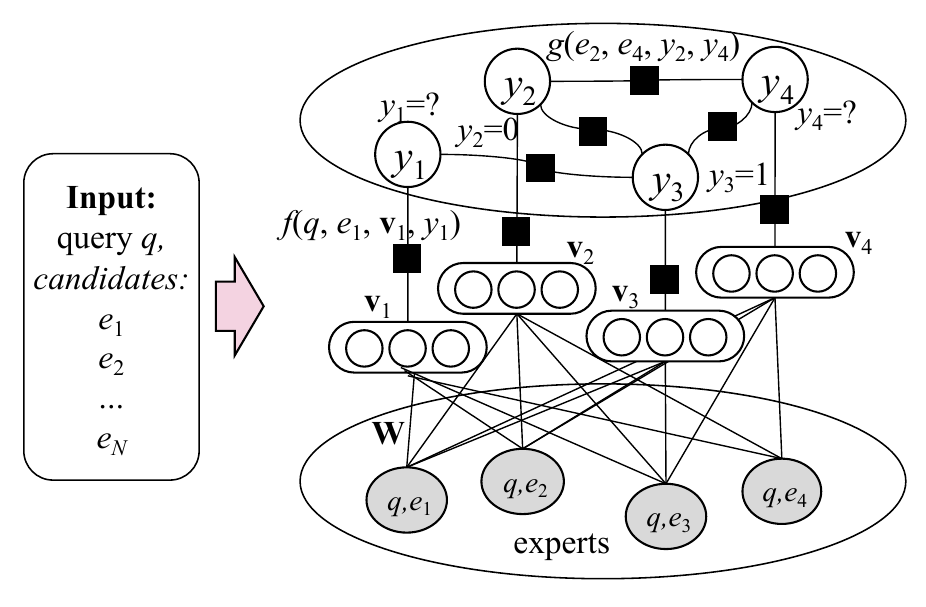}
	}
	\vspace{-0.17in} 
	\caption{
		Graphical representation of the WeakFG model. {\small Variable $y_i$ indicates whether expert $e_i$ declines the invitation; $\textbf{v}_i$ indicates the embedding for expert $e_i$; $f(q, e_i, \textbf{v}_i, y_i)$ and $g(e_i, e_j, y_i, y_j)$ represent the local factor function defined for expert $e_i$ on query $q$, and the correlation factor function defined between experts $e_i$ and $e_j$. } \label{fig:model}}
	\vspace{-0.05in}
\end{figure}

An example is illustrated in Figure~\ref{fig:model}. Given a query $q$ and candidate experts $\{e_1, e_2, e_3, e_4\}$, we first generate the embedding-based representation for each expert, i.e., $\{\textbf{v}_1, \textbf{v}_2, \textbf{v}_3, \textbf{v}_4\}$. A factor graph is then constructed by defining  local factor function $f(q, e_i, \textbf{v}_i, y_i)$ to capture  local attributes of each query-expert pair, and  correlation factor function $g(e_i, e_j, y_i, y_j)$ to capture potential correlation between different experts. Variable $y_i$ indicates whether expert $e_i$ accepts (=0) the invitation or declines (=1) it.
For example, in Figure~\ref{fig:model}, expert $e_3$ has a correlation with each other expert.
For training the factor graph model, we have two labeled instances ($y_2=0$ and $y_3=1$), thus our goal is to infer the other two unlabeled instances ($y_1$ and $y_4$).

It should be noted that each observation node $e_i$ in WeakFG actually encodes information related to both query $q$ and $e_i$. Essentially, the local factor function $f(q, e_i, \textbf{v}_i, y_i)$ aims to model the expertise matching degree, while the correlation factor function aims to \textit{propagate} the labeled information in the graph in order to help infer acceptance probabilities of  unlabeled instances.
Now we introduce how we calculate the expertise matching degree, and  explain how we instantiate the factor functions in WeakFG.
%

\vpara{Expertise Matching.}
We  define local factor functions to capture the expertise matching degree between expert and query.
Given a query $q$ and an expert $e$, we calculate the expertise matching score $R(q, e)$. 
The matching score can be defined based on various methods.
For example, one can use the keyword based model, e.g., vector-space model or language model~\cite{Zhai:01SIGIR}. 
To further capture the \textit{semantic} information, we can use
the recently developed embedding method~\cite{mikolov2013distributed}, which has shown promising results in many fields such as natural language processing, image processing, and speech recognition.

More specifically, 
a query $q$ can be considered as a document, and an expert $e$ can be considered as a document set $\{d_k\}, k\in\{1,\cdots,N_e\}$, where each  $d_k$ represents a document authored by $e$, e.g., an answer provided by $e$ or a paper published by $e$.
We define the expertise matching score $R(q, e)$ as the 
similarity between query document $q$ and the document set $\{d_k\}$, 
\beq{
	R(q, e)=\bigoplus_{k\in\{1,\cdots,N_e\}}\mathrm{Sim}(q, d_k)
	\label{eqn:q2e}
}
\noindent where $\bigoplus$ is an aggregation function, e.g., $max$ or $avg$.

We use embedding to calculate the similarity between query $q$ and document $d_k$, i.e., $\mathrm{Sim}(q, d_k)$.
To do this, we first learn a numeric vector representation (i.e., word embedding) for each word. We use Word2Vec with Skip-gram~\cite{mikolov2013distributed} to learn the word embeddings.
Word2Vec is a shallow neural network architecture  consisting of an input layer, a projection layer, and an output layer.
The training objective is to use an input word $w$ to predict surrounding words in its context. 
After the embedding training, we obtain an embedding vector $\mathbf{v}_w$  for each word $w$.
Then we estimate $\mathrm{Sim}(q, d_k)$ using two embedding-based algorithms: WMD and D2V. 

\hide{
	\begin{figure}[t]
		\centering
		\includegraphics[width=3in]{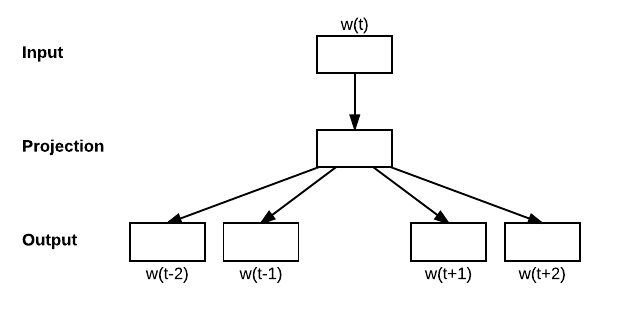}
		\caption{\label{skipgram}The Skip-gram model architecture.}
	\end{figure}
}



{\bf Word Mover's Distance (WMD).} WMD is a distance function~\cite{kusner2015word}. 
Specifically, each document $d$ can be represented as a normalized bag-of-words (nBOW) vector $\boldsymbol{\tau}\in \mathbb{R}^n$, with each element $\boldsymbol{\tau}_i=\frac{\mathcal{N}^i_d}{{\sum_{j=1}^{n}\mathcal{N}^j_d}}$ standing for the weight of word $i$ and  $\mathcal{N}^i_d$ indicating the occurrence times of word $i$ in the document. 
The similarity between documents $q$ and $d$ can be defined by the following linear program,
\beq{
	\small
	\begin{split}
		& \mathrm{Sim}_{\mathrm{WMD}}(q, d)= -\min_{\mathbf{W}_{ij}\geq 0} \sum_{i=1}^{n} \sum_{j=1}^{n} \mathbf{W}_{ij} \|\mathbf{v}_i-\mathbf{v}_j\|_2\\
		& \text{s.t. }  \sum_{j=1}^{n} \mathbf{W}_{ij} = \boldsymbol{\tau}_i^q, \  
		\sum_{i=1}^{n} \mathbf{W}_{ij} =\boldsymbol{\tau}_j ^d, \ \forall i, j \in \{ 1,\dots, n\}
	\end{split}
	\label{eqn:wmd}
	\normalsize
}
\noindent where the linear program aims to find an optimal matching between  words in $q$ and those in $d$; $\boldsymbol{\tau}^q$ and $\boldsymbol{\tau}^d$ are the nBOW vectors for $q$ and $d$. Mathematically,  each word $w_i$ in $q$ can be matched to any word $w_j$ in $d$ with the weight $\mathbf{W}_{ij}$. 
The distance between two specific words is defined as Euclidean distance $\|\cdot\|_2$ of their embeddings $\mathbf{v}_i$ and $\mathbf{v}_j$. 
We use the negative value of the distance as the similarity.



{\bf Document Vector (D2V).} 
D2V aims to directly learn an embedding vector for each document.
The document vectors can be learned by predicting all words in the document~\cite{le2014distributed}.
In this paper, we use the distributed bag of words idea. Specifically, given a document, we use stochastic gradient descent to learn the embedding vector, where in each iteration we sample a random word from the context and predict the word with the embedding vector. For the prediction, we  use the softmax function:
$$p(w|d)=\frac{\exp\left( {\mathbf{v}_{w} }^\top \mathbf{\hat{v}}_{d} \right)}{\sum_{w'=1}^{n} \exp\left( {\mathbf{v}_{w'}}^\top \mathbf{\hat{v}}_{d} \right)}$$

\noindent where  $\mathbf{\hat{v}}_{d}$ indicates the  document vector and $\mathbf{v}_w$ indicates the  word vector.
After learning the document vectors, the similarity between documents $q$ and $d$ can be defined as:
\beq{
	\mathrm{Sim}_{\mathrm{D2V}}(q, d)=\frac{{\mathbf{\hat{v}}_{q}}  {^\top} \mathbf{\hat{v}}_{d}}{\|\mathbf{\hat{v}}_{q}\|_2 \cdot \|\mathbf{\hat{v}}_{d}\|_2}
	\label{eqn:d2v}
}


\hide{
 for representing document semantics, such as language models and topic models. 
We also present two more elaborate expertise matching metrics in the next subsection.
At the same time, other features can be instantiated in different ways to reflect our prior knowledge or intuitions for different applications. They can be defined as either binary or  real-valued.
In our work, we define the feature functions ($\boldsymbol{\psi}$) which can be divided into the following two categories:
}

\vpara{WeakFG  Formalization.} 
Based on the learned embeddings (by either WMD or D2V), 
we  estimate the expertise matching scores between query and experts, and then construct the WeakFG model by defining 
 Local factor function and Correlation factor function.

\textbf{Local factor function:} It captures any attributes for each query-expert pair, including expertise matching score between the query and the expert, and any other attributes associated with  them. We define local factor  as
\beq{
f(q, e_i, \textbf{v}_i, y_i) 
= \exp \left\{ {\boldsymbol{\alpha}_{y_i}} ^\top \boldsymbol{\psi}(q, \textbf{v}_i, e_i) \right\}
}

\noindent where $\boldsymbol{\psi}(.)$ is a set of feature functions defined for $q$ and $e_i$. For example, a potential function $\psi(.)$ can be  defined as the expertise matching score $R(q, e_i)$ (Eq. (\ref{eqn:q2e})).  $\boldsymbol{\alpha}=(\boldsymbol{\alpha}_0^\top,\boldsymbol{\alpha}_1^\top)^\top$ is the weight vector of the defined features.

We can define domain-specific statistics as feature functions for each candidate expert. 
For example, in the QA expert finding task, we can define the number of questions answered by the expert as a feature, while in the paper-reviewer recommendation task, we can define features based on the expert's $h$-index, position, publication number, and citation number. 
The feature functions can be either binary or real-valued. 

\textbf{Correlation factor function:} 
The correlation between experts can be used to help infer experts' acceptance probability. It is modeled as correlation factor functions among latent variables $\{y_i\}$.
Formally, it can be also defined as:
\beq{
g(e_i, e_j, y_i, y_j) 
=  \exp \{ {\boldsymbol{\beta}_{y_i, y_j}} ^\top \boldsymbol{\phi}(e_i, e_j) \} 
}

\noindent where $\boldsymbol{\phi}(.)$ is a set of indicator functions to model whether specific correlation exists between experts $e_i$ and $e_j$; $\boldsymbol{\beta}$ indicates the weights of different factors.

In different applications, there are different types of correlations between experts.
For example, in QA expert finding, we can define correlation as co-reply,  corresponding to the situation when two users have replied to the same question. In paper-reviewer recommendation, we can consider co-author relationship.
However, as shown in Figure~\ref{fig:delinecorr}, such kinds of correlations might be still insufficient. 
We further define some other social identity~\cite{tajfel2004social} based correlations.
Social identity theory introduced a way to explain intergroup behavior, in particular to predict certain intergroup behavior on the basis of the perceived group status difference.
For example, in the QA expert finding task, users of the same position (e.g., CFO) may tend to decline the same question, e.g., about specific technique.
%
 For correlation factors, we use binary functions, i.e., $\phi_l(e_i, e_j)=1$ if and only if a specific correlation exists.

\vpara{Objective Function.}
Let $G$ denote all the observation variables, including the experts and their relations, and $Y=\{y_1,\dots,y_N\}$ denote the latent variables. 
We define the joint probability of latent variables conditioned on the observations 
by integrating the two factor functions: 
\beq{
\small
\begin{split}
& P(Y|G)  = \frac{1}{Z} \prod_{e_i} f(q,e_i, \textbf{v}_i, y_i) \prod_{(e_i, e_j)} g(e_i, e_j, y_i, y_j)  \\ 
& = \frac{1}{Z}\exp ( \sum_{e_i} {\boldsymbol{\alpha}_{y_i}} ^\top \boldsymbol{\psi}(q, \textbf{v}_i, e_i) + \sum_{(e_i, e_j)} {\boldsymbol{\beta}_{y_i, y_j}} ^\top \boldsymbol{\phi}(e_i, e_j) )
\end{split}
}

\noindent where $Z$ is a normalization factor to ensure {\small$\sum_Y P(Y|G)=1$}; 
and $\boldsymbol{\theta} = \{ \boldsymbol{\alpha}, \boldsymbol{\beta} \}$ 
are the model parameters to estimate. 

\subsection{Model Learning}

We now discuss how to estimate the model parameters $\boldsymbol{\theta} = \{ \boldsymbol{\alpha}, \boldsymbol{\beta} \}$ in WeakFG. 
Our goal is to find a parameter configuration $\boldsymbol{\theta}$ from a given training dataset $(\tilde{Y},G)$, such that the log-likelihood objective function $L(\boldsymbol{\theta})=\log P(\tilde{Y}|G)$ can be maximized, i.e.,
\beq{
 \boldsymbol{\theta}^* = \arg \max_{\boldsymbol{\theta}} \log P(\tilde{Y}|G) 
}

The optimization can be solved using a gradient ascent algorithm. As an example, we derive the gradient of $\boldsymbol{\alpha}_0$ (the gradients of other parameters can be derived analogously):
\beq{
\frac{\partial L(\boldsymbol{\theta})}{\partial \boldsymbol{\alpha}_0} 
= \sum_{e_i} \boldsymbol{\psi}(q, \textbf{v}_i, e_i) \cdot \left[\mathbb{1}(\tilde{y}_i=0) - P(y_i=0|G)\right]
\label{gradient}
}

\noindent  where $\mathbb{1}(\tilde{y}_i=0)$ is an indicator function which equals to $1$ if the label $\tilde{y}_i=0$, and $0$ otherwise.
The above gradient is intractable due to the difficulty to determine the marginal probability $P(y_i|G)$. 
There are several approximation methods.
In our work, we choose Loop Belief Propagation (LBP)~\cite{yedidia2000generalized}.
We first derive a factor graph from the original graph $G$, representing the factorization of the likelihood $P(Y|G)$. In each iteration, LBP performs message passing on the factor graph based on sum-product algorithm \cite{kschischang2001factor}, and compute the approximate marginal distributions. 
Then we compute the gradients and update the parameters with  learning rate $\eta$.


\vpara{Prediction.}
Given the input network $G$ and the parameters $\boldsymbol{\theta}$, the prediction task is to find the most likely configuration $Y_q$ for a given query $q$, i.e., 
$Y_q=\arg \max_{Y} P(Y|G).$

For prediction, we use the max-sum algorithm to find the values of $Y_q$ that maximize the likelihood. The max-sum algorithm is similar to the sum-product algorithm, except for calculating the message according to $max$ instead of $sum$ in message passing functions.

\vpara{Candidate Generation.}
Finally, we discuss a bit about the candidate generation, as it does matter in real applications.
When looking for experts for a query, we would first select  candidate experts by a coarse-level matching algorithm. This is necessary to reduce the computational cost.
Specifically, given a query $q$, we use all words in the query to select candidate experts.
We use language model (LM) to retrieve relevant experts from $V$.  LM interprets the relevance between a document and a query word as a generative probability $p(w|d)$. For  details, please refer to~\cite{Zhai:01SIGIR}.
\section{Experiments}
\label{sec:exp} 

\hide{
To empirically evaluate the effectiveness of the proposed model, we conduct experiments on three different datasets: QA-Expert, Paper-Reviewer, and Topic-Expert. Moreover, we use an online reviewer recommendation system to demonstrate the effectiveness of our model.
}
We performed both offline and online evaluations to demonstrate the effectiveness of the proposed model.
The code for this paper is publicly available.\footnote{\url{https://www.aminer.cn/match_expert}}



\begin{table*}[t]
\centering
\caption{\label{table:res} Performance comparison of different methods.}
\vspace{-0.12in}
\begin{tabular}{m{0.4in}|m{0.3in}|*{5}{c}|*{5}{c}}
\hline
\multicolumn{2}{c|}{} & \multicolumn{5}{c|}{\textbf{QA-Expert}} & \multicolumn{5}{c}{\textbf{Paper-Reviewer}} 
\\\cline{3-12}
\multicolumn{2}{c|}{Method} 
& \multicolumn{1}{m{0.3in}}{P@1} & \multicolumn{1}{m{0.3in}}{P@3} & \multicolumn{1}{m{0.3in}}{P@5} & \multicolumn{1}{m{0.3in}}{MAP} & \multicolumn{1}{m{0.3in}|}{Rprec} 
& \multicolumn{1}{m{0.3in}}{P@3} & \multicolumn{1}{m{0.3in}}{P@5} & \multicolumn{1}{m{0.3in}}{P@10} & \multicolumn{1}{m{0.3in}}{MAP} & \multicolumn{1}{m{0.3in}}{Rprec}
\\\hline
\multirow{2}*{LM}
&max 
&36.8 &34.9 &33.2 &57.4 &55.7
&66.5 &61.6 &54.7 &76.7 &67.2
\\
& avg 
&37.5 &36.8 &33.2 &57.9 &55.6
&69.9 &63.3 &54.9 &78.2 &68.3
\\
\hline
\multirow{2}*{BM25}
&max 
&33.6 &33.6 &33.3 &55.1 &53.5
&67.1 &61.8 &54.8 &77.0 &67.1
\\
& avg 
&41.4 &34.6 &33.3 &59.1 &56.5
&70.2 &63.9 &55.1 &78.1 &68.4
\\
\hline
\multirow{2}*{LDA}
&max 
&37.5 &35.3 &33.2 &56.7 &53.9
&63.0 &59.3 &54.1 &72.8 &65.6
\\
&avg
&32.9 &33.8 &32.9 &53.6 &51.4
&66.3 &62.6 &55.2 &75.8 &67.9
\\
\hline
\multirow{2}*{WMD}
&max 
&40.8 &36.4 &34.0 &59.5 &56.2
&67.3 &61.5 &54.6 &76.3 &67.2
\\
&avg
&38.2 &38.2 &34.1 &58.8 &55.6
&64.2 &59.3 &53.0 &73.7 &63.5
\\
\hline
\multirow{2}*{D2V}
&max 
&36.2 &34.4 &33.2 &57.3 &55.6
&69.6 &62.5 &55.1 &77.7 &68.3
\\
&avg
&29.6 &32.7 &33.3 &53.9 &53.0
&\textbf{72.3} &65.4 &56.2 &80.5 &71.3
\\
\hline
\multicolumn{2}{c|}{RankSVM}
&49.8 &38.5 &33.7 &65.5 &63.1
&70.6 &64.8 &56.0 &81.0 &72.7
\\

\multicolumn{2}{c|}{WeakFG}
&\textbf{52.8} 	&\textbf{39.6} &\textbf{34.1} &\textbf{67.4} &\textbf{64.5}
&71.9 &\textbf{65.5} &\textbf{56.3} &\textbf{82.0} &\textbf{73.8}
\\\hline
\end{tabular}
\vspace{-0.07in}
\end{table*}

\subsection{Experimental Setup}
 
\vpara{Datasets.}
We conducted experiments on two datasets. 

\textbf{QA-Expert:} this dataset is from an online international QA challenge$^\text{2}$.
~~It consists of 87,988 users and 47,656 questions with 43,913 invitations and 9,561 responses.
To reduce noise, we select those questions with at least 5 invited users and 1 response in our experiment. This results in 182 questions and 599 users. 
In this dataset, we aim to evaluate how can our method improve the performance of 
predicting users' responses to answer invitations in the online QA system.

\textbf{Paper-Reviewer:} this dataset comes from 
the reviewer bidding of a major CS conference in 2016. It contains 935 submissions 
 and 440 Program Committee members. PC members placed bids for the papers
 they are interested to review. 
For each paper, we consider PC's bids as a positive response to the invitation, and randomly sample the same number of PCs   who do not bid for this paper as negative. 

In each  dataset, we randomly select 60\% as the training data and  the remaining as  test data. 
We perform experiments on each dataset for 10 times and report the average results.


\vpara{Comparison Methods.}
We compare the following methods: 

\textbf{Language Model (LM)} \cite{Zhai:01SIGIR}:  We employ language model
to calculate the document similarity. We compare $max$ and $avg$ to derive $R(q,e)$ in the experiments.



\textbf{BM25} \cite{robertson2009probabilistic}: A widely used ranking function in information retrieval. The similarity score is defined as
\beq{\nonumber
\mathrm{Sim}(q,d) = \sum_{w\in q} \mathit{IDF}(w)\cdot\frac{\mathcal{N}_d^w \cdot(k_1+1)}{\mathcal{N}_d^w +k_1\cdot(1-b+b\cdot\frac{\mathcal{N}_d}{\lambda})}
}
\noindent where $\mathit{IDF}(\cdot)$ is inverse document frequency; $\mathcal{N}_d^w$ is word $w$'s frequency in document $d$, and $\mathcal{N}_d$ is the length of $d$. We set $k=2$, $b=0.75$, and $\lambda$ as the average document length.

\textbf{Latent Dirichlet Allocation (LDA)} \cite{blei2003latent}: 
We use LDA to extract the topic vectors of papers/questions, and compute the cosine similarity to get the similarity. We empirically set the number of  topics to $30$.

\textbf{Word Mover's Distance (WMD)} \cite{kusner2015word}: The similarity measure is defined in Eq.~(\ref{eqn:wmd}).

\textbf{Document Vector (D2V)} \cite{le2014distributed}: The similarity measure is defined in Eq.~(\ref{eqn:d2v}).

\textbf{RankSVM}~\cite{joachims2002optimizing}: We consider a learning to rank approach with the same feature set as WeakFG. 
For SVM, we use
 SVM-Rank implemented in SVM-Light (https://svmlight.joachims.org).  RankSVM also uses the accept/reject feedback to learn the prediction model.

\textbf{WeakFG}: The proposed method, which trains a WeakFG model to rank the experts.
For example, in QA-Expert, 
as the co-reply relationship is very sparse,
we define correlation as ``the number of both-interested topics $\geq 6$'' (each user has about 10 interested topics).

\vpara{Evaluation Measures.}
We consider the problem as a ranking problem and evaluate different methods 
by Precision of the top-$N$ results (P@$N$),  
Mean Average Precision (MAP), and R-prec~\cite{Buckley:04}.
%

\subsection{Experimental Results}

Table \ref{table:res} lists the performance of all  methods on the datasets. 
Overall, the proposed WeakFG clearly outperforms all the comparison methods by 1.5-10.7\% in terms of MAP.

\vpara{How embeddings help?}
Taking a closer look, we observed that 
the embedding-based methods (WMD and D2V) perform better than the other methods (LM and BM25) using only keywords. LDA can be also considered as an embedding-like algorithm by using topics. However its performance is not as good as expected. Both RankSVM and WeakFG leverage the embedding results and improve the prediction performance.

\vpara{How correlations help?}
One advantage of our model is the incorporation of correlations to deal with the sparsity problem. As the comparison, RankSVM  considers users' responses, but does not consider correlations.
WeakFG incorporates various correlations to propagate the responses throughout the graph model, and thus achieves better performance.

\hide{
\begin{table}[!hbp]
\begin{tabular}{|c|c|c|c|c|}
\hline
\hline
lable 1-1 & label 1-2 & label 1-3 & label 1 -4 & label 1-5 \\
\hline
label 2-1 & label 2-2 & label 3-3 & label 4-4 & label 5-5 \\
\hline
\multirow{2}{*}{Multi-Row} & \multicolumn{2}{c}{Multi-Column} & \multicolumn{2}{|c|}{\multirow{2}{*}{Multi-Row and Col}} \\
\cline{2-3}
& column-1 & column-2 & \multicolumn{2}{|c|}{}\\
\hline
\end{tabular}
\caption{My first table}
\end{table}

\section{Conclusions}
\label{sec:conclusion} 


In this paper, we study an interesting problem of matching  experts in online community.
We propose a weakly supervised factor graph (WeakFG) model to formalize the entire problem into a principled framework. 
WeakFG  combines expertise matching and correlations between experts using local and correlation factor functions. 
Our experiments on two different genres of datasets validate the effectiveness  of the proposed model.
We also deployed an online system to further demonstrate the advantages of the proposed model.

\vpara{Acknowledgements.} The work is supported by the
National Basic Research Program of China (2014CB340506),
the Open Project  of the 
State Key Laboratory of Mathematical Engineering and Advanced Computing (2016A07),
%
and 
the Royal Society-Newton Advanced Fellowship. Jie Tang is the corresponding author.

\newpage
\small
\bibliographystyle{named}
\bibliography{references}

\end{document}